\documentclass[A4paper,conference]{IEEEtran}
\IEEEoverridecommandlockouts
\usepackage{cite}
\usepackage{amsmath,amssymb,amsfonts}
\usepackage{algorithmic}
\usepackage{graphicx}
\usepackage{textcomp}
\usepackage{xcolor}
\usepackage{soul}
\usepackage{url}
\usepackage{booktabs}
\usepackage{multirow}
\usepackage{makecell}
\usepackage{subcaption}
\usepackage{mdframed}
\usepackage{xurl}
\usepackage[hidelinks]{hyperref}
\def\BibTeX{{\rm B\kern-.05em{\sc i\kern-.025em b}\kern-.08em
    T\kern-.1667em\lower.7ex\hbox{E}\kern-.125emX}}

\makeatletter
\def\ps@IEEEtitlepagestyle{%
  \def\@oddfoot{}%
  \def\@evenfoot{}%
}
\newcommand{\linebreakand}{%
  \end{@IEEEauthorhalign}
  \hfill\mbox{}\par
  \mbox{}\hfill\begin{@IEEEauthorhalign}
}
\makeatother
    
\begin{document}

\title{Iterative Hybrid Discrete-Continuous Viewpoint Planning for UAV Photogrammetry\\
 \thanks{This work was conducted as part of the joint industry-academic project \textit{Automated Aerial Capture (AAC)} between \href{https://www.stargatestudios.com.mt/}{Stargate Studios Malta} and the University of Malta (\href{https://www.um.edu.mt/research/dawl/}{Dawl AI Lab}), financed by Xjenza Malta through the Technology Extension Support Programme (TESP).}
}

\author{\IEEEauthorblockN{Alan Grech}
\IEEEauthorblockA{
University of Malta\\
Msida, Malta \\
alan.grech@um.edu.mt}
\and
\IEEEauthorblockN{Daniel Pisani}
\IEEEauthorblockA{
University of Malta\\
Msida, Malta \\
daniel.o.pisani@um.edu.mt }
\and
\IEEEauthorblockN{Andre Grima}
\IEEEauthorblockA{
Stargate Studios Malta\\
Mosta, Malta \\
andre.grima@stargatestudios.com.mt}
\linebreakand
\IEEEauthorblockN{Carl James Debono}
\IEEEauthorblockA{
University of Malta\\
Msida, Malta \\
c.debono@ieee.org}
\and
\IEEEauthorblockN{Saviour Formosa}
\IEEEauthorblockA{
University of Malta\\
Msida, Malta \\
saviour.formosa@um.edu.mt}
\and
\IEEEauthorblockN{Dylan Seychell}
\IEEEauthorblockA{
University of Malta\\
Msida, Malta \\
dylan.seychell@um.edu.mt}
}

\maketitle

\begin{abstract}
Unmanned aerial vehicle (UAV) photogrammetry requires camera networks that provide sufficient surface coverage, image overlap, parallax, and resolution, yet conventional flight patterns are often poorly adapted to scene geometry resulting in local reconstruction errors. This paper proposes an iterative hybrid discrete-continuous viewpoint planning method for targeted UAV photogrammetry from a proxy reconstruction. The method scores sampled surface points using photogrammetric heuristics based on frontality, imaging distance, parallax, and multi-view observation count, while also evaluating the full viewpoint set in terms of visibility, pairwise overlap, and graph connectivity. Candidate viewpoints are generated around weakly observed regions, refined using clustered Covariance matrix adaptation evolution strategy (CMA-ES) optimisation, and removed when redundant. The final flight path combines close-range detail viewpoints with wider model-coverage viewpoints, balancing local reconstruction quality with global image-network robustness. Evaluation on three synthetic scenes shows that the proposed method improves both reconstruction accuracy and completeness compared with prior UAV path-planning methods.
\end{abstract}


\begin{mdframed}[
    linewidth=0.4pt,
    roundcorner=4pt,
    innertopmargin=6pt,
    innerbottommargin=6pt,
    innerleftmargin=8pt,
    innerrightmargin=8pt
]
\centering
\small
\textit{This paper has been accepted for publication in the\\
14th IEEE European Conference on Visual Information Processing (EUVIP 2026).}
\end{mdframed}

\section{Introduction}
Unmanned aerial vehicle (UAV) photogrammetry is widely used for reconstructing buildings, infrastructure, cultural heritage sites, and outdoor environments. However, reconstruction quality depends strongly on the acquired camera network, which must provide sufficient coverage, overlap, parallax, and resolution while avoiding poor viewing angles and occlusions~\cite{Smith2018,Roberts2017,Pataki2025}. Conventional trajectories, such as nadir lawnmower flights, are simple to execute but are not adapted to scene geometry or local view-quality requirements~\cite{Zhang2020,Zhang2024}. As a result, complex structures may remain incomplete or imprecise when surfaces are observed from unsuitable angles, excessive distances, or too few overlapping images~\cite{Roberts2017,Hepp2019}.

This paper proposes an iterative photogrammetry-aware UAV viewpoint planning method. Following prior proxy-based planning approaches, an initial exploratory flight produces a coarse reconstruction, which is then used to plan a targeted second flight. The planner evaluates sampled surface points using frontality, imaging distance, parallax, and multi-view observation count, while also scoring the full viewpoint set using visibility, pairwise overlap, and graph connectivity.

The planner follows a hybrid discrete-continuous strategy. Candidate viewpoints are generated around weakly observed surface regions, refined through clustered continuous optimisation, and removed when redundant. The final trajectory combines close-range detail viewpoints for local reconstruction quality with wider model-coverage viewpoints for contextual observations and image-network connectivity.

The main contributions of this paper are:
\begin{itemize}
    \item A reconstruction objective that combines local photogrammetric quality with global image-network properties.
    \item A hybrid discrete-continuous planner that adds, optimises, and removes viewpoints iteratively.
    \item A two-level viewpoint design combining close-range detail views with wider model-coverage views.
\end{itemize}

\section{Related Work}

Camera placement directly affects both the Structure-from-Motion (SfM) and Multi-View Stereo (MVS) stages of photogrammetric reconstruction. SfM requires sufficient image overlap for feature matching and enough viewpoint separation for well-conditioned triangulation, while MVS depends on consistent multi-view correspondences, suitable viewing angles, and redundant observations. Low overlap, weak texture visibility, excessive viewpoint changes, small baselines, occlusions, and poor surface frontality can therefore reduce registration reliability, completeness, and depth accuracy~\cite{Pataki2025,Rumpler2011}. View planning methods address these issues by selecting camera poses that improve reconstruction conditions, and can broadly be divided into online next-best-view (NBV) approaches and model-based planning methods.

NBV methods incrementally select the next camera pose from the current reconstruction state. Huang et al.~\cite{Huang2018} reconstruct a coarse model during flight and add views for under-sampled regions, while learning-based approaches such as Hepp et al.~\cite{Hepp2018}, Guédon et al.~\cite{Guedon2023}, and Chen et al.~\cite{Chen2024} predict viewpoint utility from occupancy, uncertainty, RGB input, or learned scene representations. These methods are valuable when no prior model is available, but are mainly designed for online exploration and coverage gain rather than complete photogrammetric image-network optimisation.

Model-based methods are more closely aligned with planned UAV photogrammetry, where an initial proxy model from a preliminary flight or external data source can be used to design object-aware flight paths. Smith et al.~\cite{Smith2018} propose a continuous optimisation approach for aerial urban reconstruction, using reconstructability heuristics to refine sampled camera positions and orientations. Roberts et al.~\cite{Roberts2017} formulate aerial scanning as a submodular orienteering problem, selecting views that maximise coverage and view diversity under a travel budget. Plan3D by Hepp et al.~\cite{Hepp2019} similarly adopts a proxy-based formulation, jointly considering viewpoint selection, trajectory constraints, and free-space feasibility for aerial MVS. Zhang et al.~\cite{Zhang2021} further formulate planning as a continuous aerial path planning problem using a view information field and RRT-based planning, allowing view utility, flight quality, and path smoothness to be considered jointly. Together, these methods show the benefit of planning from an intermediate scene representation and, in some cases, moving beyond fixed candidate selection toward continuous or hybrid optimisation.

Other methods focus on identifying and repairing weak reconstruction regions. Zhang et al.~\cite{Zhang2020,Zhang2024} use quality indicators such as incompleteness, curvature, viewing angle, ground sampling distance (GSD), and precision estimates to guide additional viewpoint planning. These indicators are derived by comparing the Dense Image Matching point cloud with a sampled point cloud on the proxy model. Yan et al.~\cite{Yan2021} similarly distinguish between local viewpoints for low-quality regions and global viewpoints for overall scene coverage. This distinction reflects an important property of UAV photogrammetry: close-range views can improve local detail and resolution, while wider views may be needed to preserve coverage, context, and image-network stability.

\section{Methodology}
\subsection{Overview}
\begin{figure}
    \centering
    \includegraphics[width=1\linewidth]{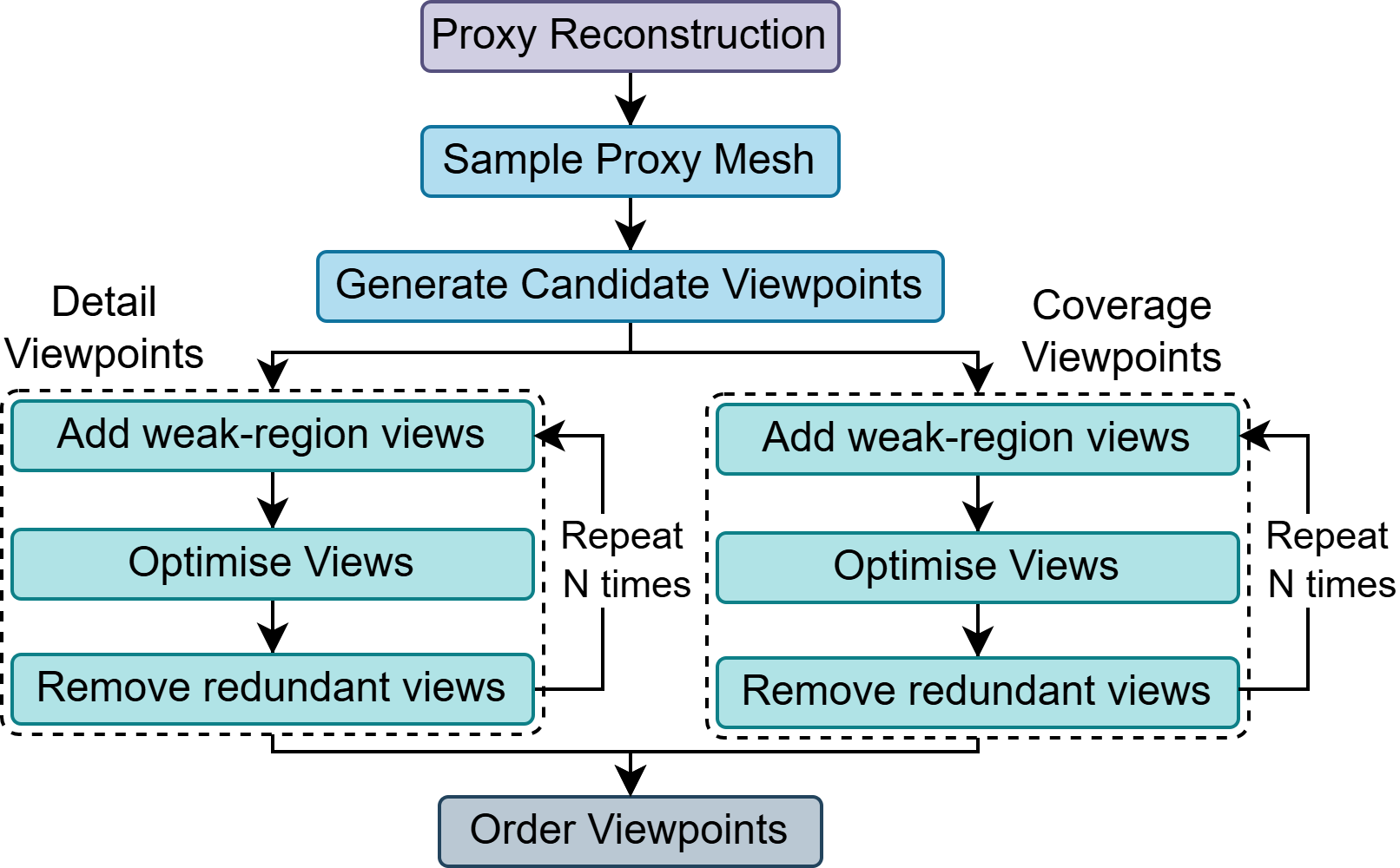}
    \caption{Overview of the proposed planner. A proxy reconstruction is sampled to generate detail and coverage viewpoints, which are iteratively added, optimised with CMA-ES, culled, and ordered into a flight path.}
    \label{fig:overview}
\end{figure}

The proposed method, illustrated in Figure~\ref{fig:overview}, generates a photogrammetry-aware UAV flight path from a proxy reconstruction. A two-stage acquisition process is assumed, where an initial exploratory flight, such as a nadir lawnmower trajectory, is first used to reconstruct a proxy mesh. This mesh is then used to plan a second, targeted flight that improves surface coverage, local reconstruction quality, image overlap, and viewpoint connectivity.


The pipeline consists of candidate generation, viewpoint addition, clustered continuous optimisation, culling, model-coverage viewpoint generation, and trajectory ordering. The final set contains close-range detail viewpoints for weak regions and wider model-coverage viewpoints for global connectivity.

\subsection{Reconstruction Heuristics}
\subsubsection{Point-Level Reconstruction Heuristics}

The point-level reconstruction heuristic estimates the expected reconstruction quality of a sampled mesh point from its visible viewpoints. It combines surface frontality, imaging distance, parallax, and multi-view observation count:

\begin{equation}
    R_{point}(p, V_p) = S_{\text{front}} \cdot S_{\text{distance}} \cdot S_{\text{parallax}} \cdot S_{\text{count}}
\end{equation}
where $p$ is a sampled mesh point, $V_p$ is the set of viewpoints that observe it and $S_{\text{front}}$, $S_{\text{distance}}$, $S_{\text{parallax}}$ and $S_{\text{count}}$ are the frontality, distance, parallax and observation count scores respectively.

The frontality score favours observations close to the surface normal, reducing perspective distortion and improving image support for matching and texture reconstruction. For a point $p$ with normal $n$, the best frontal observation is used:

\begin{equation}
    S_{\text{front}}(p, n, V_p)
    =
    \max_{v \in V_p}
    \max\left(0,\,
    n \cdot
    \frac{v_{pos} - p}{\|v_{pos} - p\|}
    \right)
\end{equation}

The distance score favours observations near the target distance derived from the desired GSD. Unlike Smith et al.~\cite{Smith2018}, arbitrarily close viewpoints are not rewarded, rather observations receive maximum score within the target range and are linearly penalised until they exceed a tolerated margin controlled by $d_{\text{margin}}$.

The parallax score encourages sufficient angular diversity while penalising excessive viewpoint separation. The angular spread $\sigma_\alpha$ of the viewing directions is passed through the logistic penalties $w_1(\sigma_\alpha)$ and $w_3(\sigma_\alpha)$ defined by Smith et al.~\cite{Smith2018}:

\begin{equation}
    S_{\text{parallax}}(p, V_p) = w_1(\sigma_\alpha)w_3(\sigma_\alpha)
\end{equation}
The angular spread is computed as:
\begin{equation}
    \sigma_{\alpha}^{2} = \frac{\sum_{v \in V_p} \arccos(u \cdot \bar{u})^2}{|V_p| + \epsilon}, \quad \bar{u} = \frac{\sum_{v \in V_p} u_v}{\left\|\sum_{v \in V_p} u_v\right\|}
\end{equation}
where $u_v$ denotes the unit viewing direction from viewpoint $v$ to point $p$.

The observation count score ensures that each point is seen from enough viewpoints:

\begin{equation}
    S_{\text{count}}(V_p)
    =
    \min\left(1, \frac{|V_p|}{k_{\text{count}}}\right)
\end{equation}
where $k_{\text{count}} = 3$ is used in practice.

\subsubsection{Viewpoint-Set Reconstruction Heuristics}
Point-level scores do not fully describe the suitability of the complete image network, since a viewpoint set may provide good local observations while still forming disconnected clusters or containing redundant images. The viewpoint-set heuristic therefore rewards visibility, pairwise overlap, and graph connectivity:

\begin{equation}
    R_{viewpoint}(P, V)
    =
    \frac{1}{3}
    \left(
    S_{\text{visibility}}
    +
    S_{\text{overlap}}
    +
    S_{\text{conn}}
    \right)
\end{equation}
where $P$ is the set of sampled mesh points, $V$ is the viewpoint set and $S_{\text{visibility}}$, $S_{\text{overlap}}$ and $S_{\text{conn}}$ are the visibility, overlap and connectivity scores respectively.

The visibility score measures the average number of sampled points observed by each viewpoint using a binary visibility matrix $A$, where $A_{ij}=1$ if point $p_j$ is visible from viewpoint $v_i$:

\begin{equation}
    S_{\text{visibility}}
    =
    \frac{1}{|V||P|}
    \sum_{i=1}^{|V|}
    \sum_{j=1}^{|P|}
    A_{ij}
\end{equation}

The viewpoint overlap score encourages each viewpoint to share sufficient scene content with neighbouring viewpoints. For each pair, the overlap ratio $r_{ik}$ is the number of commonly visible points divided by the smaller individual visible-point count. This ratio is mapped to a pairwise score $f(r_{ik})$ using a clipped linear ramp between $r_{\min}$ and $r_{\max}$. The final contribution of each viewpoint is the mean of its top $k$ pairwise overlap scores:

\begin{equation}
    S_{\text{overlap}}
    =
    \frac{1}{|V|}
    \sum_{i=1}^{|V|}
    \frac{1}{k}
    \sum_{s \in T_k(i)} s,
    \quad
    T_k(i)=\text{TopK}_{j \ne i}(f(r_{ij}))
\end{equation}

To encourage a connected image network, a graph $G$ is constructed with viewpoints as nodes and edges between pairs whose overlap score exceeds $\tau_{\mathrm{edge}}$. The connectivity score is the fraction of viewpoints in the largest connected component:

\begin{equation}
    S_{\mathrm{conn}}
    =
    \frac{1}{|V|}
    \max_{C \in \mathcal{C}(G)}
    |C|
\end{equation}
where $\mathcal{C}(G)$ denotes the set of connected components of the graph $G$. \\
The global reconstruction score combines point-level quality with the viewpoint-set score:

\begin{equation}
    \begin{aligned}
        R(P, V) =
        &\sum_{p \in P}
        R_{\text{point}}
        \left(
        p,
        \{v_i \in V \mid p \text{ is visible from } v_i\}
        \right) \\
        &+ |P|R_{\text{viewpoint}}(P, V)
    \end{aligned}
\end{equation}

\subsection{Candidate Viewpoint Generation}
\label{subsec:viewgen}
Candidate viewpoints are generated around surface points that are insufficiently observed or have low reconstruction scores. At each iteration, the planner selects the lowest-scoring mesh point that has not already been selected and generates a candidate group around it at the target imaging distance.

For each selected point, approximately frontal viewpoints are sampled around the surface normal. Additional side viewpoints are sampled at a specified angular offset from each frontal direction. Each candidate is oriented towards the target point subject to camera and gimbal constraints, and invalid candidates are removed using visibility checks. The frontal viewpoint most closely aligned with the point normal is selected first, followed by a subset of side viewpoints chosen to be angularly spread apart. This provides local multi-view diversity while keeping the group compact. If no valid group can be found, the point is excluded from further candidate generation.

\subsection{Viewpoint Addition}
The initial viewpoint addition stage ensures that each sampled point is observed by at least the minimum required number of viewpoints. Weak-region repair is then performed by revisiting points whose reconstruction score remains below a threshold. New candidate groups are generated using the same local sampling strategy and are accepted only if they improve the global objective. This prevents viewpoints from being added when they improve a weak point locally but reduce the overall quality or connectivity of the image network.

\subsection{Clustered Viewpoint Optimisation}
The discrete generation stage produces an initial feasible viewpoint set, but individual poses may remain suboptimal. The method therefore refines viewpoints using the Covariance Matrix Adaptation Evolution Strategy (CMA-ES). To reduce dimensionality, viewpoints are partitioned into spatial clusters and each cluster is optimised separately. Each viewpoint is parameterised by its 3D position and viewing direction, represented by pitch and yaw. Roll is excluded because consumer UAV cameras are typically stabilised about the roll axis by the gimbal. Candidate poses are constrained by local position bounds, gimbal limits, and a minimum distance from the mesh.

CMA-ES is used because the reconstruction objective is non-convex, non-differentiable, and discontinuous due to discrete visibility changes and the coupled effects of distance, parallax, overlap, and connectivity. It is therefore more suitable than gradient-based optimisation and preferable to the Nelder--Mead method used by Smith et al.~\cite{Smith2018}. Although Nelder--Mead is gradient-free~\cite{Nelder1965,Lagarias1998}, it has limited convergence guarantees and can fail even on smooth low-dimensional functions~\cite{Lagarias1998,McKinnon1998}. CMA-ES is better suited to this rugged, non-separable problem because it adapts both the sampling covariance and global step size to exploit correlated search directions~\cite{Hansen2001,Hansen2016}.
~\cite{Hansen2016}.

\subsection{Viewpoint Culling}
After generation and optimisation, viewpoints that contribute little additional value are removed. Viewpoints that observe no sampled points are discarded directly. Highly overlapping viewpoint pairs are then considered as redundancy candidates. For each pair, the method evaluates the effect of removing either viewpoint and accepts the removal only if the global score is not reduced beyond a permitted threshold. This reduces unnecessary image acquisition while preserving reconstruction quality and viewpoint-set connectivity.

\subsection{Model-Coverage Viewpoints}
In addition to close-up detail viewpoints, the method generates model-coverage viewpoints to provide broader scene context and improve image-network robustness. Unlike detail viewpoints, which are generated around local surface points at the target GSD distance, model-coverage viewpoints are positioned to frame larger portions of the proxy model. Their score prioritises the number of visible sampled points, with a small preference for closer valid observations.

The model-coverage pass follows the same add--optimise--cull structure as the detail-viewpoint pass, but uses a coverage-oriented scoring function and a larger effective distance margin. The final trajectory therefore combines local high-detail observations with wider contextual views that support global reconstruction consistency.

\subsection{Trajectory Planning}
The planning stages produce an unordered set of camera viewpoints. To convert this set into an executable UAV trajectory, the viewpoints are ordered by solving a Travelling Salesman Problem. This produces an efficient traversal order without changing the selected photogrammetric viewpoints.

\section{Evaluation}
\subsection{Experimental Setup}
Evaluation was performed on three models: Tu Duc's Tomb\footnote{\textit{Tu Duc's Tomb}, available at \url{https://skfb.ly/o7pvI}. Accessed: 30 May 2026.}, Church of St. Sophia\footnote{\textit{Church of St. Sophia}, available at \url{https://skfb.ly/6YIMB}. Accessed: 30 May 2026.}, and Mexico City Metropolitan Cathedral\footnote{\textit{Mexico City Metropolitan Cathedral}, available at \url{https://skfb.ly/6RvxJ}. Accessed: 30 May 2026.}. Images were rendered in Unity 6000.3.2f1. For each model, a proxy reconstruction was first generated from approximately 200 images captured in a lawnmower grid pattern, with viewpoints directed towards the object centre. The proxy was reconstructed using OpenMVG and OpenMVS, while the final reconstruction was produced using RealityScan, creating the final reconstruction at normal detail quality.

The proposed method was compared with Smith et al.~\cite{Smith2018} and Yan et al.~\cite{Yan2021} using the accuracy and completeness metrics defined by Smith et al.~\cite{Smith2018}. Accuracy is reported as the distance below which a specified percentile of reconstructed points lie from the ground-truth mesh, while completeness is the percentage of ground-truth points lying within a fixed distance threshold of the reconstruction.

Smith et al.'s method was run at a front and side overlap of 80\% and a minimum distance of 5 metres. Yan et al.'s method was given 4000 sampled points of each scene. The settings for both methods were selected according to the default parameter values provided by their respective authors. Our method sampled 2048 points and ran 4 optimisation rounds for the close-up detail viewpoints and 3 optimisation rounds for the coverage viewpoints. A viewpoint cluster size of 12 was used for both cases. Images were captured at a resolution of 2048$\times$1536px, providing sufficient image detail while keeping reconstruction times moderate.

\subsection{Results}

\begin{figure}[t]
    \centering

    \begin{subfigure}{0.48\columnwidth}
        \centering
        \includegraphics[width=\linewidth]{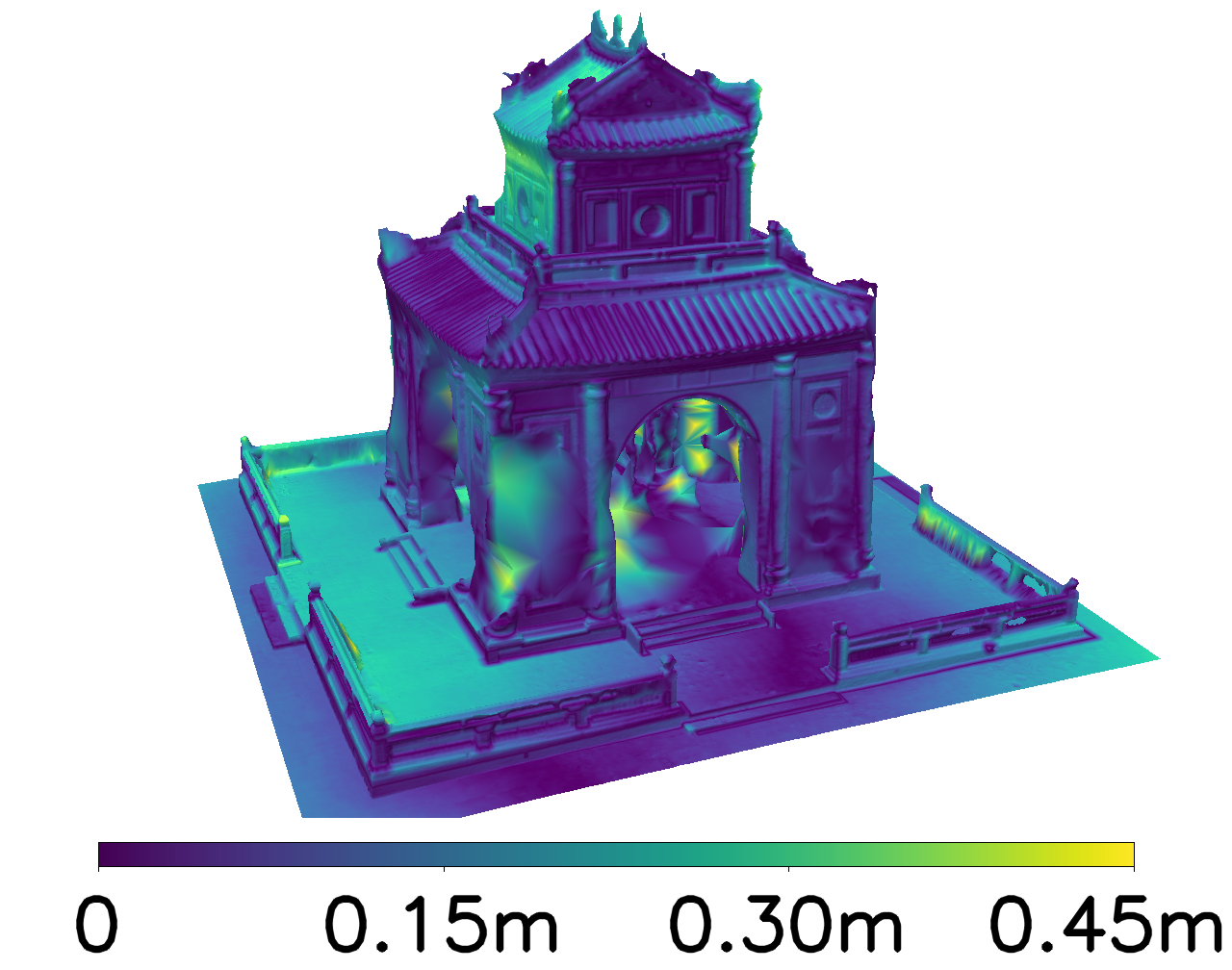}
        \caption{Smith et al. \cite{Smith2018}}
        \label{fig:smith_tomb_error}
    \end{subfigure}
    \hfill
    \begin{subfigure}{0.48\columnwidth}
        \centering
        \includegraphics[width=\linewidth]{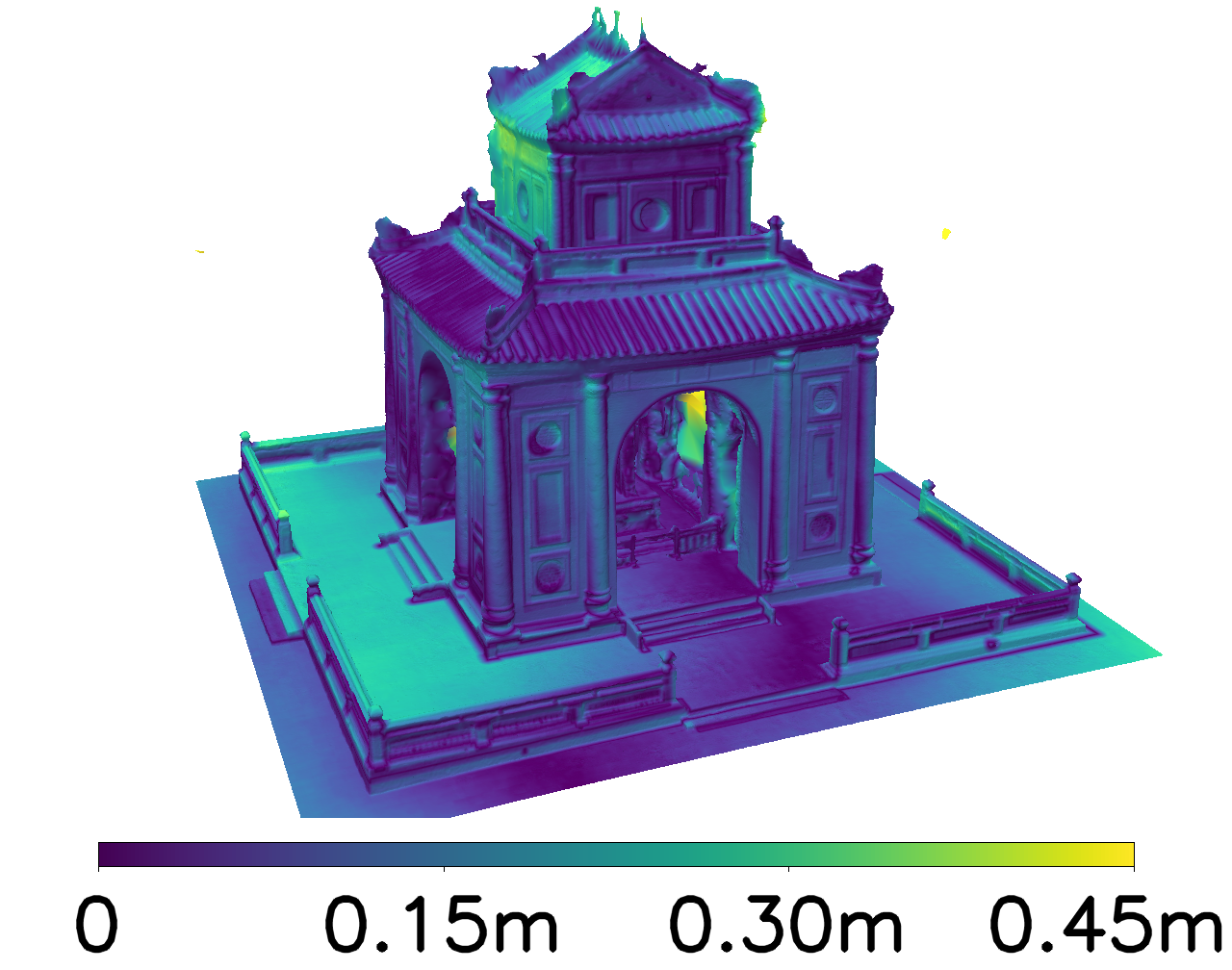}
        \caption{Yan et al. \cite{Yan2021}}
        \label{fig:yan_tomb_error}
    \end{subfigure}

    \vspace{2mm}

    \begin{subfigure}{0.48\columnwidth}
        \centering
        \includegraphics[width=\linewidth]{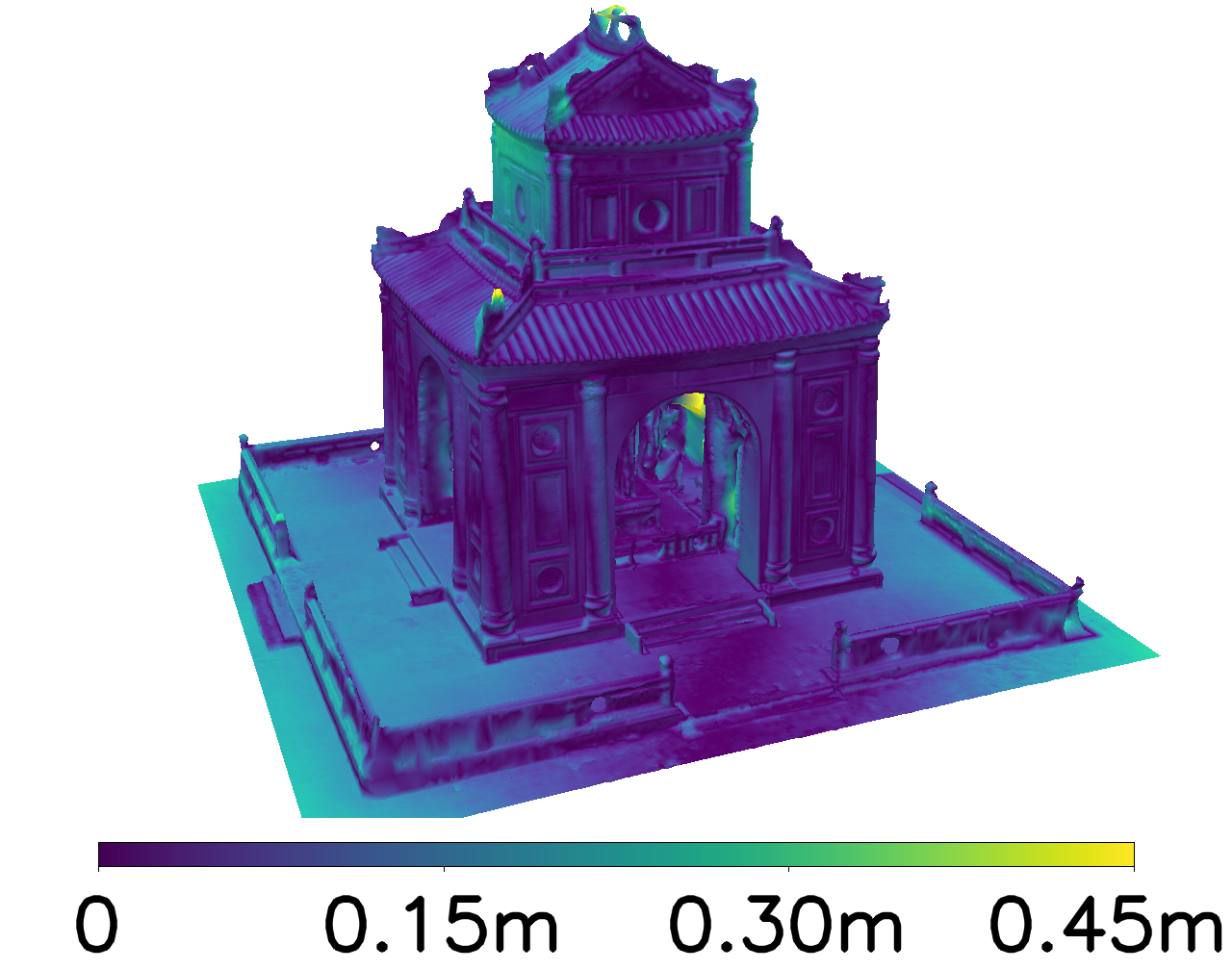}
        \caption{Our method}
        \label{fig:our_tomb_error}
    \end{subfigure}

    \caption{Reconstruction error visualisations for Smith et al. (a), Yan et al. (b) and Our method (c).}
    \label{fig:reconstruction_errors}
\end{figure}

Table~\ref{tab:recon_results} shows that the proposed method generally improves reconstruction accuracy across the evaluated scenes. The clearest gains are observed for Tomb and Church, where it obtains the best results at both error percentiles. For Cathedral, the method achieves the lowest 90th-percentile error, although Yan et al.~\cite{Yan2021} performs better at the 95th percentile, suggesting that some larger errors remain in more challenging regions. This is consistent with the Tomb error visualisations in Figure~\ref{fig:reconstruction_errors}, where our method produces fewer severe error regions around the facade and entrance than the two baselines.

\begin{figure}[t]
    \centering

    \begin{subfigure}{0.49\columnwidth}
        \centering
        \includegraphics[width=\linewidth]{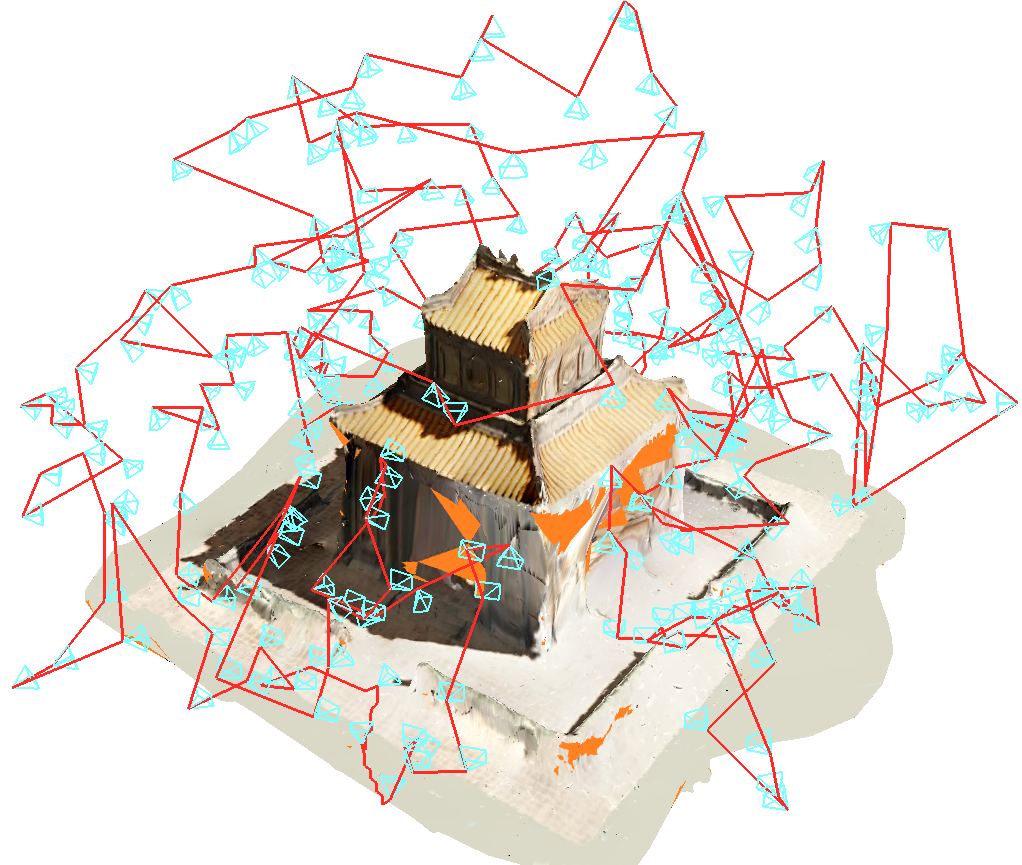}
        \caption{Tomb}
        \label{fig:tomb_flight_path}
    \end{subfigure}
    \hfill
    \begin{subfigure}{0.49\columnwidth}
        \centering
        \includegraphics[width=\linewidth]{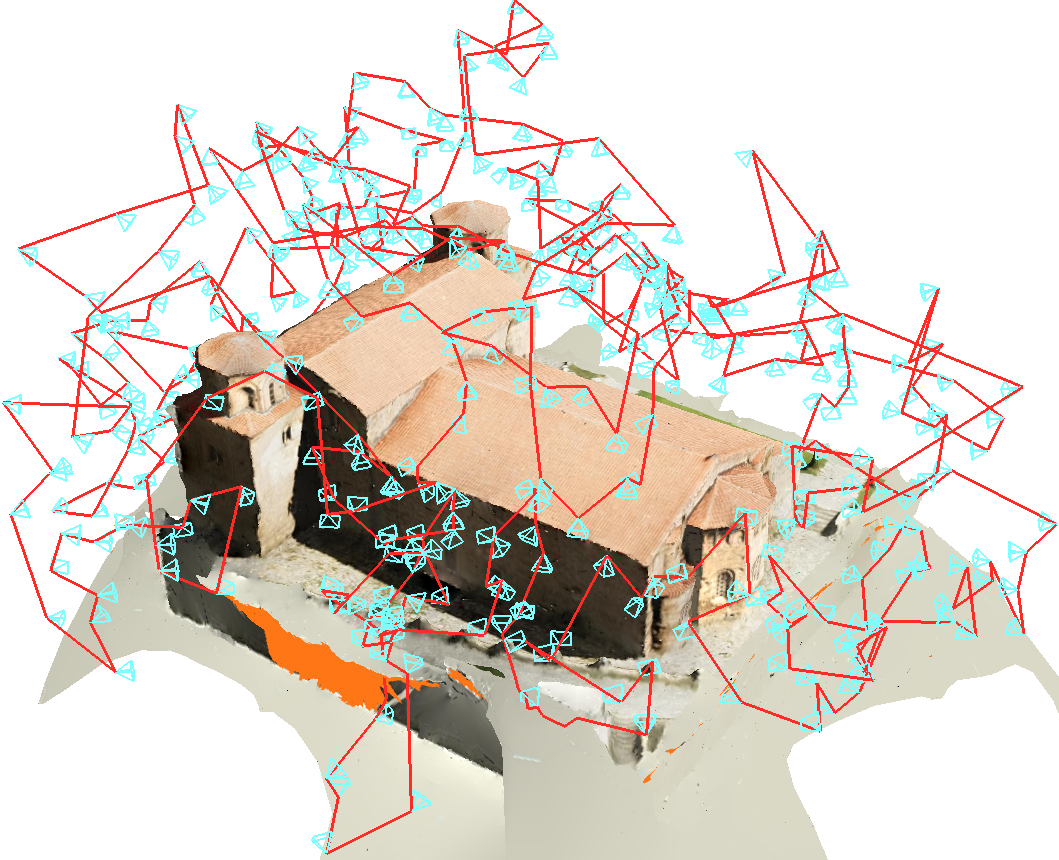}
        \caption{Church}
        \label{fig:church_flight_path}
    \end{subfigure}
    
    \begin{subfigure}{0.49\columnwidth}
        \centering
        \includegraphics[width=\linewidth]{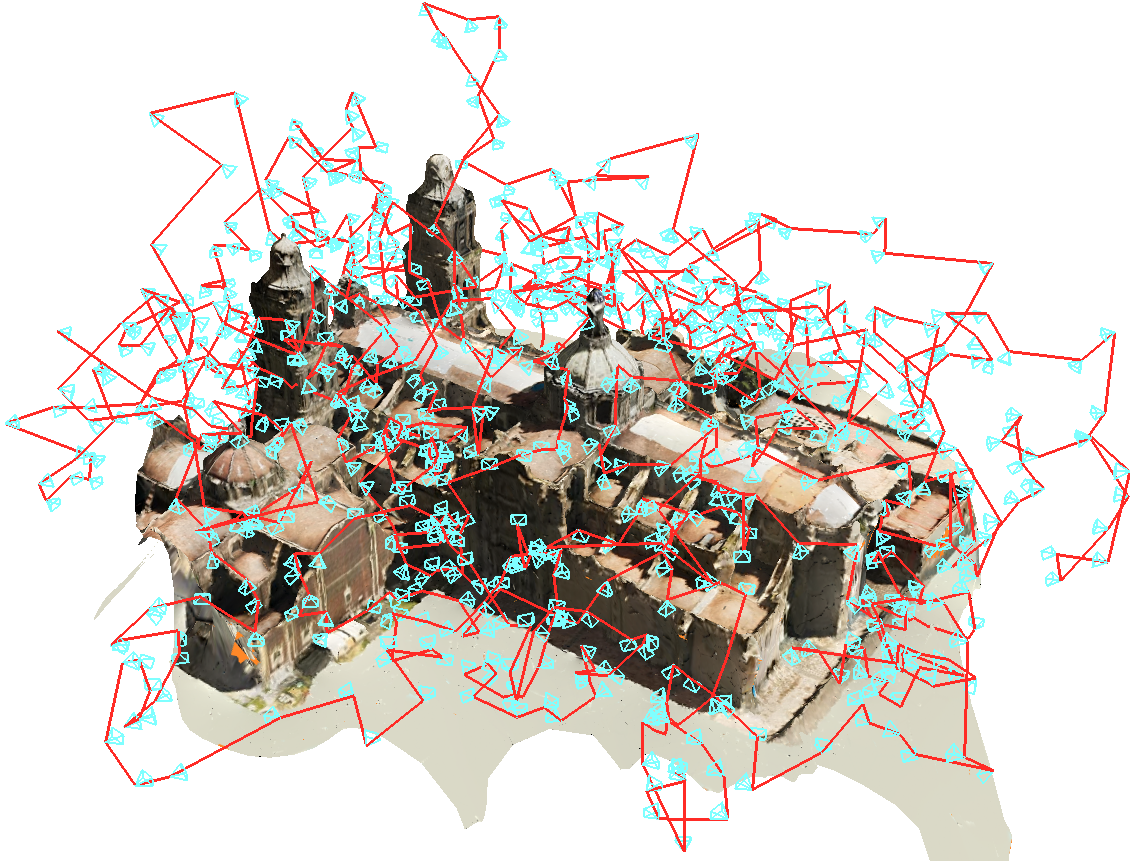}
        \caption{Cathedral}
        \label{fig:cathedral_flight_path}
    \end{subfigure}

    \caption{Flight paths generated by the proposed method for the three scenes.}
    \label{fig:flight_paths}
\end{figure}

The completeness results show a similar trend. The proposed method achieves the highest completeness at the 0.075~m threshold across all scenes and performs best at the stricter 0.050~m threshold for Tomb and Cathedral. Yan et al.~\cite{Yan2021} slightly outperforms it on Church at 0.050~m, indicating that the improvement in surface coverage is partly threshold- and scene-dependent. The flight paths in Figure~\ref{fig:flight_paths} show viewpoints distributed around each structure and ensuring coverage on complex regions, contributing to the more uniform error distribution in the Tomb scene. Although completeness improves in most cases, the absolute values remain limited, particularly for Tomb and Cathedral. This may be partly due to the ground-truth meshes being reconstructions of real-world artefacts with high-frequency surface noise, as well as internal or recessed regions being absent from the proxy model.

\begin{table}[t]
\centering
\caption{Reconstruction accuracy and completeness results for the three scenes.}
\label{tab:recon_results}
\setlength{\tabcolsep}{4pt}
\begin{tabular}{llrrrrr}
\toprule
Scene & Method 
& \makecell{Error\\90\% (m)} 
& \makecell{Error\\95\% (m)} 
& \makecell{Comp.\\0.050m (\%)} 
& \makecell{Comp.\\0.075m (\%)} \\
\midrule
\multirow{3}{*}{Tomb}
& \cite{Smith2018} & 0.266 & 0.389 & 18.92 & 27.47 \\
& \cite{Yan2021} & 0.238 & 0.281 & 24.29 & 35.61 \\
& Ours  & \textbf{0.196} & \textbf{0.247} & \textbf{31.56} & \textbf{42.76} \\
\midrule
\multirow{3}{*}{Church}
& \cite{Smith2018} & 0.115 & 0.139 & 51.97 & 61.66 \\
& \cite{Yan2021} & 0.106 & 0.145 & \textbf{59.39} & 69.92 \\
& Ours  & \textbf{0.093} & \textbf{0.118} & 57.23 & \textbf{70.95} \\
\midrule
\multirow{3}{*}{Cathedral}
& \cite{Smith2018} & 0.351 & 0.587 & 22.56 & 31.86 \\
& \cite{Yan2021} & 0.206 &  \textbf{0.276} & 33.73 & 47.44 \\
& Ours  & \textbf{0.198} & 0.366 & \textbf{36.88} & \textbf{49.87} \\
\bottomrule
\end{tabular}
\end{table}

Table~\ref{tab:align_results} shows that these improvements are achieved without a substantial increase in alignment cost. Our method uses fewer images than Yan et al.~\cite{Yan2021} in all scenes and fewer than Smith et al.~\cite{Smith2018} for Tomb and Church. Its alignment time is also substantially lower than Yan et al., while remaining comparable to Smith et al., despite generally producing higher reconstruction quality.

\begin{table}[t]
\centering
\caption{Number of input images and alignment time for each flight path method.}
\label{tab:align_results}
\begin{tabular}{llrrrrr}
\toprule
Scene & Method & Number of Images & \makecell{ Alignment\\time (minutes)} \\
\midrule
\multirow{3}{*}{Tomb}
& Smith et al.~\cite{Smith2018} & 322 & 0.77 \\
& Yan et al.~\cite{Yan2021} & 3,795 & 174 \\
& Ours  & 291 & 0.78 \\
\midrule
\multirow{3}{*}{Church}
& Smith et al.~\cite{Smith2018} & 874 & 1.61 \\
& Yan et al.~\cite{Yan2021} & 3,441 & 19.18 \\
& Ours  & 419 & 1.1 \\
\midrule
\multirow{3}{*}{Cathedral}
& Smith et al.~\cite{Smith2018} & 920 & 2.53 \\
& Yan et al.~\cite{Yan2021} & 3,373 & 26.92 \\
& Ours  & 828 & 2.83 \\
\bottomrule
\end{tabular}
\end{table}

Overall, the results indicate that the proposed flight path provides a favourable trade-off between reconstruction quality and acquisition efficiency. The viewpoint distributions in Figure~\ref{fig:flight_paths} support this interpretation, as the paths remain relatively sparse while still targeting structurally important regions. This allows the method to improve accuracy and completeness in most cases while using a relatively small number of images and maintaining short alignment times.

\section{Conclusion}
This paper presented an iterative hybrid discrete-continuous viewpoint planning method for UAV photogrammetry. The method uses a proxy reconstruction to identify weakly observed surface points and generate additional viewpoints that improve local reconstruction quality while preserving image-network visibility, overlap, and connectivity. By combining close-range detail views with wider model-coverage views, the planner balances local accuracy with global reconstruction robustness.

The evaluation shows that the proposed method improves both reconstruction accuracy and completeness in most cases, achieving the best 90th-percentile error and the highest 0.075 m completeness across all scenes, while using substantially fewer images than Yan et al. and maintaining alignment times comparable to Smith et al. This indicates a better trade-off between reconstruction quality and acquisition cost.

A key limitation of the proposed method is its dependence on the quality of the proxy reconstruction. Since the second flight is planned from the proxy mesh, regions that are missing or poorly reconstructed in the proxy, such as the internal geometry observed in the Tomb scene, may not be sufficiently prioritised during viewpoint generation. Future work will therefore investigate the use of the proxy reconstruction images, in addition to the proxy mesh, to detect details that were not captured in the proxy geometry, and will incorporate flight-time and flight-quality terms into the CMA-ES objective so that reconstruction quality and practical UAV trajectory efficiency can be jointly optimised.

\bibliographystyle{IEEEtran}
\bibliography{references}

@misc{Huang2018,
	title = {Active {Image}-based {Modeling} with a {Toy} {Drone}},
	url = {http://arxiv.org/abs/1705.01010},
	doi = {10.48550/arXiv.1705.01010},
	urldate = {2025-08-20},
	publisher = {arXiv},
	author = {Huang, Rui and Zou, Danping and Vaughan, Richard and Tan, Ping},
	month = mar,
	year = {2018},
	note = {arXiv:1705.01010 [cs]},
}

@misc{Chen2024,
	title = {{GenNBV}: {Generalizable} {Next}-{Best}-{View} {Policy} for {Active} {3D} {Reconstruction}},
	shorttitle = {{GenNBV}},
	url = {http://arxiv.org/abs/2402.16174},
	doi = {10.48550/arXiv.2402.16174},
	urldate = {2025-08-20},
	publisher = {arXiv},
	author = {Chen, Xiao and Li, Quanyi and Wang, Tai and Xue, Tianfan and Pang, Jiangmiao},
	month = jul,
	year = {2024},
	note = {arXiv:2402.16174 [cs]},
}

@inproceedings{Guedon2023,
	title = {{MACARONS}: {Mapping} and {Coverage} {Anticipation} with {RGB} {Online} {Self}-{Supervision}},
	issn = {2575-7075},
	shorttitle = {{MACARONS}},
	url = {https://ieeexplore.ieee.org/document/10204865},
	doi = {10.1109/CVPR52729.2023.00097},
	urldate = {2025-12-16},
	booktitle = {2023 {IEEE}/{CVF} {Conference} on {Computer} {Vision} and {Pattern} {Recognition} ({CVPR})},
	author = {Guédon, Antoine and Monnier, Tom and Monasse, Pascal and Lepetit, Vincent},
	month = jun,
	year = {2023},
	pages = {940--951},
}

@inproceedings{Hepp2018,
	address = {Cham},
	title = {Learn-to-{Score}: {Efficient} {3D} {Scene} {Exploration} by {Predicting} {View} {Utility}},
	isbn = {978-3-030-01267-0},
	shorttitle = {Learn-to-{Score}},
	doi = {10.1007/978-3-030-01267-0_27},
	language = {en},
	booktitle = {Computer {Vision} – {ECCV} 2018},
	publisher = {Springer International Publishing},
	author = {Hepp, Benjamin and Dey, Debadeepta and Sinha, Sudipta N. and Kapoor, Ashish and Joshi, Neel and Hilliges, Otmar},
	editor = {Ferrari, Vittorio and Hebert, Martial and Sminchisescu, Cristian and Weiss, Yair},
	year = {2018},
	pages = {455--472},
}

@inproceedings{Roberts2017,
	title = {Submodular {Trajectory} {Optimization} for {Aerial} {3D} {Scanning}},
	issn = {2380-7504},
	url = {https://ieeexplore-ieee-org.ejournals.um.edu.mt/document/8237831},
	doi = {10.1109/ICCV.2017.569},
	urldate = {2025-10-16},
	booktitle = {2017 {IEEE} {International} {Conference} on {Computer} {Vision} ({ICCV})},
	author = {Roberts, Mike and Shah, Shital and Dey, Debadeepta and Truong, Anh and Sinha, Sudipta and Kapoor, Ashish and Hanrahan, Pat and Joshi, Neel},
	month = oct,
	year = {2017},
	pages = {5334--5343},
}

@article{Smith2018,
	title = {Aerial path planning for urban scene reconstruction: a continuous optimization method and benchmark},
	volume = {37},
	issn = {0730-0301},
	shorttitle = {Aerial path planning for urban scene reconstruction},
	url = {https://dl.acm.org/doi/10.1145/3272127.3275010},
	doi = {10.1145/3272127.3275010},
	number = {6},
	urldate = {2025-10-16},
	journal = {ACM Trans. Graph.},
	author = {Smith, Neil and Moehrle, Nils and Goesele, Michael and Heidrich, Wolfgang},
	month = dec,
	year = {2018},
	pages = {183:1--183:15},
}

@article{Hepp2019,
	title = {{Plan3D}: {Viewpoint} and {Trajectory} {Optimization} for {Aerial} {Multi}-{View} {Stereo} {Reconstruction}},
	volume = {38},
	issn = {0730-0301, 1557-7368},
	shorttitle = {{Plan3D}},
	url = {https://dl.acm.org/doi/10.1145/3233794},
	doi = {10.1145/3233794},
	language = {en},
	number = {1},
	urldate = {2025-10-15},
	journal = {ACM Transactions on Graphics},
	author = {Hepp, Benjamin and Nießner, Matthias and Hilliges, Otmar},
	month = feb,
	year = {2019},
	pages = {1--17},
}

@article{Zhang2024,
	title = {Guided by model quality: {UAV} path planning for complete and precise {3D} reconstruction of complex buildings},
	volume = {127},
	issn = {1569-8432},
	shorttitle = {Guided by model quality},
	url = {https://www.sciencedirect.com/science/article/pii/S1569843224000219},
	doi = {10.1016/j.jag.2024.103667},
	urldate = {2025-10-07},
	journal = {International Journal of Applied Earth Observation and Geoinformation},
	author = {Zhang, Shuhang and Liu, Chun and Haala, Norbert},
	month = mar,
	year = {2024},
	pages = {103667},
}

@article{Zhang2020,
	title = {{THREE}-{DIMENSIONAL} {PATH} {PLANNING} {OF} {UAVS} {IMAGING} {FOR} {COMPLETE} {PHOTOGRAMMETRIC} {RECONSTRUCTION}},
	volume = {V-1-2020},
	issn = {2194-9042},
	url = {https://isprs-annals.copernicus.org/articles/V-1-2020/325/2020/},
	doi = {10.5194/isprs-annals-V-1-2020-325-2020},
	language = {English},
	urldate = {2025-12-05},
	journal = {ISPRS Annals of the Photogrammetry, Remote Sensing and Spatial Information Sciences},
	publisher = {Copernicus GmbH},
	author = {Zhang, S. and Liu, C. and Haala, N.},
	month = aug,
	year = {2020},
	note = {Conference Name: XXIV ISPRS Congress, Commission I (Volume V-1-2020) - 2020 edition},
	pages = {325--331},
}

@article{Zhang2021,
	title = {Continuous aerial path planning for {3D} urban scene reconstruction},
	volume = {40},
	issn = {0730-0301},
	url = {https://dl.acm.org/doi/10.1145/3478513.3480483},
	doi = {10.1145/3478513.3480483},
	number = {6},
	urldate = {2025-12-07},
	journal = {ACM Trans. Graph.},
	author = {Zhang, Han and Yao, Yucong and Xie, Ke and Fu, Chi-Wing and Zhang, Hao and Huang, Hui},
	month = dec,
	year = {2021},
	pages = {225:1--225:15},
}

@article{Yan2021,
	title = {Sampling-{Based} {Path} {Planning} for {High}-{Quality} {Aerial} {3D} {Reconstruction} of {Urban} {Scenes}},
	volume = {13},
	copyright = {http://creativecommons.org/licenses/by/3.0/},
	issn = {2072-4292},
	url = {https://www.mdpi.com/2072-4292/13/5/989},
	doi = {10.3390/rs13050989},
	language = {en},
	number = {5},
	urldate = {2026-01-14},
	journal = {Remote Sensing},
	publisher = {publisher},
	author = {Yan, Feihu and Xia, Enyong and Li, Zhaoxin and Zhou, Zhong},
	month = mar,
	year = {2021},
}

@article{Rumpler2011,
author = {Rumpler, Markus and Irschara, Arnold and Bischof, Horst},
year = {2011},
month = {01},
pages = {},
title = {Multi-View Stereo: Redundancy Benefits for 3D Reconstruction}
}

@inproceedings{Pataki2025,
	title = {{MP}-{SfM}: {Monocular} {Surface} {Priors} for {Robust} {Structure}-{From}-{Motion}},
	issn = {2575-7075},
	shorttitle = {{MP}-{SfM}},
	url = {https://ieeexplore.ieee.org/document/11094431},
	doi = {10.1109/CVPR52734.2025.02039},
	urldate = {2026-05-26},
	booktitle = {2025 {IEEE}/{CVF} {Conference} on {Computer} {Vision} and {Pattern} {Recognition} ({CVPR})},
	author = {Pataki, Zador and Sarlin, Paul-Edouard and Schönberger, Johannes L. and Pollefeys, Marc},
	month = jun,
	year = {2025},
	note = {ISSN: 2575-7075},
	pages = {21891--21901},
}

@misc{Hansen2016,
	title = {The {CMA} {Evolution} {Strategy}: {A} {Tutorial}},
	shorttitle = {The {CMA} {Evolution} {Strategy}},
	url = {https://arxiv.org/abs/1604.00772v2},
	language = {en},
	urldate = {2026-05-28},
	journal = {arXiv.org},
	author = {Hansen, Nikolaus},
	month = apr,
	year = {2016},
}

@article{Nelder1965,
	title = {A {Simplex} {Method} for {Function} {Minimization}},
	volume = {7},
	issn = {0010-4620, 1460-2067},
	url = {https://academic.oup.com/comjnl/article-lookup/doi/10.1093/comjnl/7.4.308},
	doi = {10.1093/comjnl/7.4.308},
	language = {en},
	number = {4},
	urldate = {2026-05-28},
	journal = {The Computer Journal},
	author = {Nelder, J. A. and Mead, R.},
	month = jan,
	year = {1965},
	pages = {308--313},
}

@article{Lagarias1998,
	address = {Philadelphia, United States},
	title = {Convergence {Properties} of the {Nelder}--{Mead} {Simplex} {Method} in {Low} {Dimensions}},
	volume = {9},
	copyright = {[Copyright] © 1998 Society for Industrial and Applied Mathematics},
	issn = {10526234},
	url = {https://www.proquest.com/docview/920036749/abstract/15D6D23F76B946A3PQ/1},
	doi = {10.1137/S1052623496303470},
	language = {English},
	number = {1},
	urldate = {2026-05-28},
	journal = {SIAM Journal on Optimization},
	publisher = {Society for Industrial and Applied Mathematics},
	author = {Lagarias, Jeffrey C. and Reeds, James A. and Wright, Margaret H. and Wright, Paul E.},
	year = {1998},
	note = {Num Pages: 36},
	pages = {36},
}

@article{Hansen2001,
	title = {Completely {Derandomized} {Self}-{Adaptation} in {Evolution} {Strategies}.},
	volume = {9},
	issn = {1063-6560},
	url = {https://research.ebsco.com/plink/580e4cdd-3f2d-3804-8db0-065dceda5711},
	doi = {10.1162/106365601750190398},
	language = {eng},
	number = {2},
	urldate = {2026-05-28},
	journal = {Evolutionary Computation},
	publisher = {MIT Press},
	author = {Hansen, Nikolaus and Ostermeier, Andreas},
	month = jun,
	year = {2001},
	pages = {159--195},
}

@article{McKinnon1998,
	address = {Philadelphia, United States},
	title = {Convergence of the {Nelder}--{Mead} {Simplex} {Method} to a {Nonstationary} {Point}},
	volume = {9},
	copyright = {[Copyright] © 1998 Society for Industrial and Applied Mathematics},
	issn = {10526234},
	url = {https://www.proquest.com/docview/920036271/abstract/2D1F69323B2F40B0PQ/1},
	doi = {10.1137/S1052623496303482},
	language = {English},
	number = {1},
	urldate = {2026-05-28},
	journal = {SIAM Journal on Optimization},
	publisher = {Society for Industrial and Applied Mathematics},
	author = {McKinnon, K. I. M.},
	year = {1998},
	note = {Num Pages: 11},
	pages = {11},
}

\end{document}